\documentclass{article} 
\usepackage{iclr2027_conference,times}

\usepackage{amsmath,amsfonts,bm}

\def\eqref#1{equation~\ref{#1}}

\def\1{\bm{1}}

\DeclareMathAlphabet{\mathsfit}{\encodingdefault}{\sfdefault}{m}{sl}
\SetMathAlphabet{\mathsfit}{bold}{\encodingdefault}{\sfdefault}{bx}{n}

\usepackage{hyperref}
\usepackage{url}
\usepackage{booktabs}
\usepackage{tabularx}
\usepackage{algorithm}
\usepackage{algpseudocode}
\usepackage{amssymb}
\usepackage{graphicx}
\usepackage{subcaption}

\title{Adapting Nonstationary Multi-output Gaussian Processes to Bayesian Optimization}

\author{Zikai Xie \thanks{Corresponding author.} \\
State Key Laboratory of Precision and Intelligent Chemistry\\
University of Science and Technology of China\\
Hefei, Anhui 230000, China \\
\texttt{zikaix@ustc.edu.cn}
}

\iclrfinalcopy 
\begin{document}

\maketitle

\begin{abstract}
Multi-objective Bayesian optimization (MOBO) commonly relies on independent Gaussian processes (GPs) with stationary kernels, limiting its ability to represent nonstationary structure and share information between objectives. However, expressive nonstationary GPs do not necessarily make reliable BO decisions. We study this mismatch for the multi-output low-rank nonstationary (MO-LRN) GP: strong training fit can coexist with large off-design errors and optimistic acquisition predictions. We introduce MOLRN-BO, which combines a regularized shared-spectral surrogate with objective-specific residuals, prequential mean correction and tempered covariance scaling, and Pareto-local qLogEHVI optimization with periodic global search. Experiments on 12 deterministic bi-objective benchmarks show that MOLRN-BO substantially improves upon the original MO-LRN and achieves the best average problem ranks for final normalized hypervolume and normalized inverted generational distance among nine evaluated algorithms. It also achieves the strongest adverse-tail performance while remaining competitive with the leading baselines in anytime optimization. Ablation studies further show that the shared spectral construction improves off-design prediction, the local--global decision policy improves optimization performance, and hierarchical calibration reduces systematic candidate bias. These results demonstrate that nonstationary multi-output surrogates can deliver strong and robust MOBO performance when their structure and use are explicitly adapted to the demands of sequential optimization.

\end{abstract}

\section{Introduction}

Expensive scientific and engineering optimization often requires identifying
good trade-offs among conflicting objectives with few black-box evaluations~\citep{terayama2021black,shahriari2015taking,luo2025physics}. Bayesian optimization (BO)~\citep{terayama2021black} and its multi-objective variants (MOBO)~\citep{laumanns2002bayesian} address this problem by fitting a probabilistic surrogate and selecting experiments through an acquisition function. Its success depends
on predictions at the points the acquisition function prefers, which may differ
substantially from the observed training distribution.

Methodological advances in MOBO have largely concentrated on acquisition design, including scalarization-, hypervolume-, and information-theoretic criteria~\citep{knowles2006parego,daulton2020differentiable,daulton2021parallel,hernandez2016predictive,belakaria2019max,tu2022joint}. By contrast, standard implementations typically retain independent GPs with stationary Matérn or RBF kernels, which neither explicitly share information across objectives nor represent input-dependent covariance structure. Recently, the multi-output low-rank nonstationary (MO-LRN) kernel offers a more expressive alternative through low-rank spectral factors~\citep{xu2026revisiting}. Yet, our preliminary experiments show that increasing its spectral capacity improved training fit while worsening off-design prediction. In addition, predictions at acquisition-selected candidates were biased and underdispersed. This regression–decision gap motivates our central question: how should MO-LRN be structured and deployed as a reliable surrogate for MOBO?

To address these issues, we introduce MOLRN-BO, a decision-oriented nonstationary multi-output BO framework. MOLRN-BO stabilizes the surrogate through regularized shared LRN components with per-mixture coregionalization and objective-specific ARD-RBF residuals. It then constructs a decision posterior using strictly prequential bias estimates and bounded covariance rescaling, and optimizes qLogEHVI through Pareto-centered local search with periodic global rounds. Together, these components align surrogate fitting, predictive calibration, and acquisition optimization with the requirements of sequential multi-objective decision making.

Our contributions are threefold. First, we identify and characterize the mismatch between
training fit, off-design prediction, and acquisition decisions for MO-LRN and develop, to our knowledge, the first MOBO
framework built on a harmonizable low-rank nonstationary multi-output GP.
Second, evaluations on 12 bi-objective benchmarks show that MOLRN-BO
achieves state-of-the-art final Pareto quality among the evaluated methods with the best average ranks in both normalized hypervolume and normalized IGD,
while maintaining competitive anytime performance. Third, controlled ablations quantify the effects of spectral capacity
and sharing, joint local--global search and mode-specific bias estimation,
and hierarchical bias pooling on prediction and optimization.

\section{Related Work}

\paragraph{Multi-objective surrogates and acquisitions.} Research in MOBO has primarily focused on
acquisition design. Scalarization-based methods, including ParEGO and random scalarization, reduce the vector-valued problem to a sequence of scalar subproblems~\citep{knowles2006parego,paria2020flexible}. Hypervolume-based methods, such as EHVI, qEHVI, qNEHVI, and their logarithmic variants, directly reward improvement in dominated hypervolume~\citep{emmerich2006single, daulton2020differentiable,daulton2021parallel}. Information-theoretic methods, including PESMO, MESMO, JES, and PF$^2$ES, instead select evaluations that reduce uncertainty about the Pareto set or Pareto front~\citep{hernandez2016predictive,belakaria2019max,tu2022joint}. Surrogate design has received comparatively less attention: beyond the independent stationary GPs used by standard methods, a smaller line of work considers correlated multi-task or multi-output GPs and deep GP surrogates~\citep{shah2016pareto,chowdhury2021no,hebbal2023deep}. However, nonstationary spectral multi-output surrogates remain largely unexplored in acquisition-driven MOBO. 

\paragraph{Nonstationary spectral models.}
Multi-output GPs place a joint prior over vector-valued functions through a matrix-valued kernel whose diagonal entries describe within-output covariance and whose off-diagonal entries describe cross-output dependence. Classical constructions such as the intrinsic coregionalization model and the linear
model of coregionalization obtain valid cross-covariances by mixing latent
scalar GPs~\citep{alvarez2012kernels}. Spectral alternatives instead model these dependencies in the frequency domain: MOSM represents output-specific magnitudes, delays, and phases but remains stationary~\citep{parra2017spectral}, whereas MOHSM uses a harmonizable double-frequency representation to permit input-dependent auto- and cross-covariances~\citep{altamirano2022nonstationary}. MO-LRN further combines a generalized spectral--kernel duality with a low-rank factorization of flexible matrix-valued nonstationary spectral densities, reducing parameter growth in the number of outputs from quadratic to linear~\citep{xu2026revisiting}. Previous evaluations of MO-LRN have focused on passive prediction; we examine the additional demands that arise when it is deployed inside an acquisition-driven MOBO loop. 

\paragraph{Calibration and localized search.}

Acquisition functions depend on both posterior means and posterior
uncertainties, making predictive calibration especially important under model
misspecification and adaptive sampling. Previous work has used conformal
prediction, posterior recalibration, and online residual correction to improve
predictive coverage or quantile estimates in BO
~\citep{stanton2023bayesian,deshpande2024online}. Most of these methods are
developed for single-output BO or focus primarily on marginal predictive
coverage, and calibration for correlated nonstationary surrogates under
hypervolume-based acquisition remains comparatively unexplored. On the other hand, localized BO provides a complementary approach to stabilizing optimization in
large or difficult search spaces. TuRBO maintains adaptive trust regions for
single-objective BO, whereas MORBO coordinates multiple local regions to cover
different parts of a multi-objective Pareto front
~\citep{eriksson2019scalable,daulton2022multi}. Both approaches couple
localization with local surrogate modeling and search. 

\section{Method}
\label{sec:method}

\begin{figure}[ht!]
    \centering
    \includegraphics[width=0.91\linewidth]{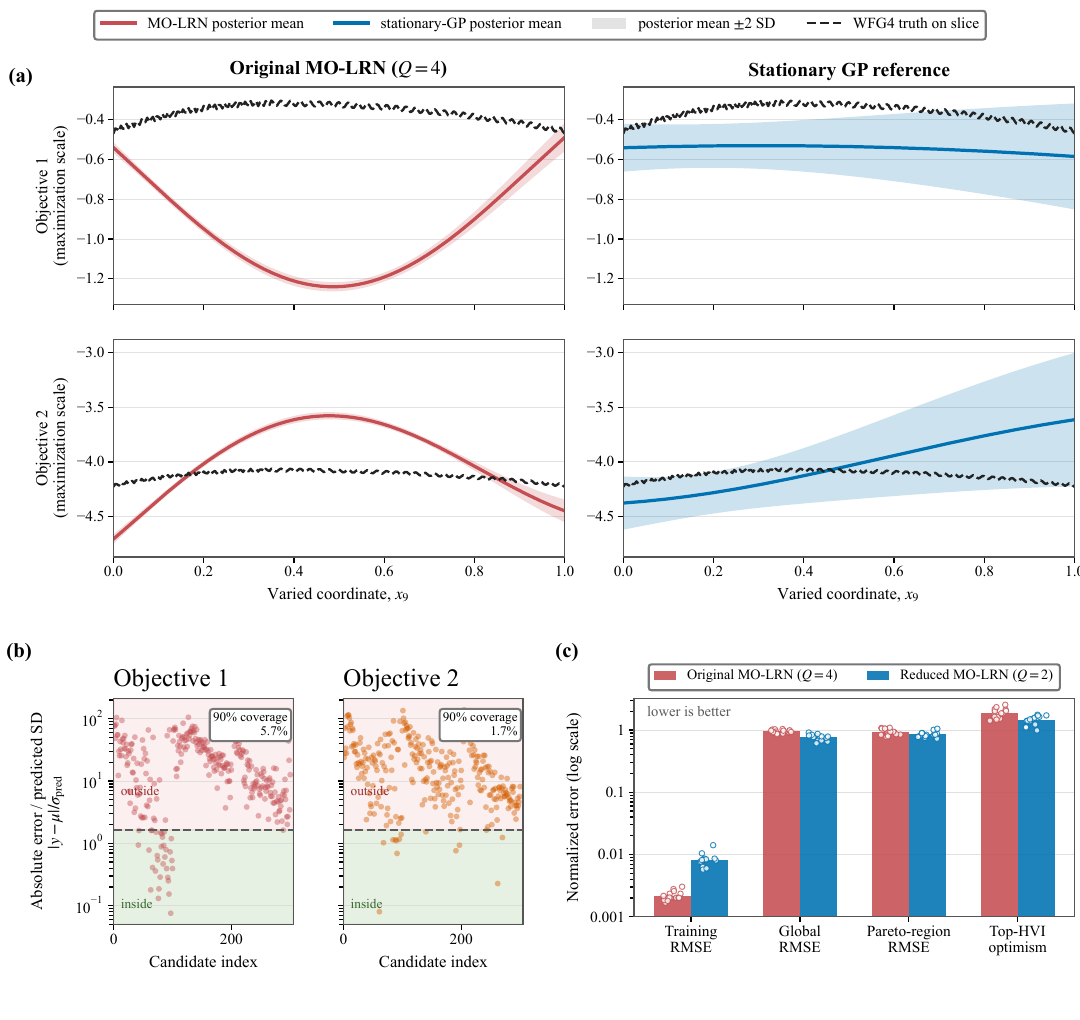}
    \caption{
Development diagnostics for the original MO-LRN surrogate on WFG4.
\textbf{(a)} One-dimensional posterior slice at $n=20$, obtained by varying $x_9$ while fixing the remaining coordinates at their training-data medians. Solid curves show posterior means, shaded regions show $\pm2$ posterior standard deviations, and dashed curves show the true objectives.
\textbf{(b)} Strictly prequential standardized prediction errors, $|y-\mu|/\sigma_{\mathrm{pred}}$, at acquisition-selected candidates over three trials. Predictions are evaluated before each candidate is added to the training set; the dashed line at 1.645 denotes the nominal two-sided 90\% predictive limit.
\textbf{(c)} Training, global, and Pareto-region RMSE together with top-HVI optimism for $Q=2$ and $Q=4$ MO-LRN models fitted to identical observations at $n=110$. Bars denote means and points denote 15 fits from three data trajectories and five restarts.
}
    \label{fig:motivation}
\end{figure}

We maximize $\mathbf f:\mathcal X\subset\mathbb R^D\to\mathbb R^M$ on a
bounded box. After $n$ evaluations, let
$\mathcal D_n=\{(\mathbf x_t,\mathbf y_t)\}_{t=1}^n$ and let
$\mathcal P_n$ be its nondominated objective vectors. Experiments use
deterministic objectives and one candidate per round. Objective-wise
standardization is recomputed on $\mathcal D_n$, with standard deviations
$\mathbf s_n$. Raw posterior moments $\boldsymbol\mu_n(\mathbf x)$ and
$\boldsymbol\Sigma_n(\mathbf x)$ below are on the original objective scale.

\paragraph{From surrogate fit to decision reliability.}

Sequential model-based optimization generally relies on the premise that a more accurate surrogate provides a better basis for subsequent optimization decisions. However, when applying the original MO-LRN surrogate to WFG4, we observed that improved regression fit did not consistently translate into a more reliable posterior at the locations relevant to BO. Figure~\ref{fig:motivation}a provides a representative example: along a one-dimensional slice, MO-LRN with higher spectral capacity produces strongly distorted posterior means together with relatively narrow uncertainty bands, despite being fitted to the same observations. Compared to a stationary RBF kernel GP reference, this imbalance between posterior mean and uncertainty is particularly problematic for acquisition-based optimization, where confident extrapolation can directly affect candidate selection. Consistent with this observation, strictly prequential predictions at acquisition-selected candidates are severely underdispersed (Fig.~\ref{fig:motivation}b). Moreover, increasing the number of spectral mixtures from $Q=2$ to $Q=4$ reduces training RMSE while worsening both off-design prediction and HVI estimation (Fig.~\ref{fig:motivation}c). 

Additional pilot diagnostics revealed related symptoms, including spectral frequencies approaching their imposed bounds, weak overlap between output-specific spectra, and a tendency for unrestricted acquisition optimization to select distant or boundary-adjacent candidates (Appendix~\ref{app:pilot_diagnostics}). These observations suggest that the difficulty is not captured by regression accuracy alone. Instead, reliable use of MO-LRN in sequential optimization requires controlling spectral capacity, calibrating the posterior specifically at decision-relevant locations, and preventing acquisition optimization from overexploiting poorly supported regions. These considerations motivate the three corresponding components of MOLRN-BO developed below.

\paragraph{Shared-spectral covariance.}
To share nonstationary structure across objectives without relying on the
overlap of independently fitted output spectra, we use the LMC-style covariance
\begin{equation}
 K_{ij}(\mathbf x,\mathbf x')=
 \sum_{q=1}^{Q}[\mathbf B_q]_{ij}
 k_q^{\mathrm{sLRN}}(\mathbf x,\mathbf x')
 +\delta_{ij}k_i^{\mathrm{RBF}}(\mathbf x,\mathbf x'),
 \qquad \mathbf B_q\succeq0.
 \label{eq:composite_kernel}
\end{equation}
Each scalar component $k_q^{\mathrm{sLRN}}$ retains the MO-LRN nonstationary
construction with spectral parameters shared across objectives. The matrix
$\mathbf B_q$ represents component-specific cross-objective dependence,
while an ARD-RBF residual $k_i^{\mathrm{RBF}}$ captures smooth variation
specific to objective $i$. We use $Q=2$ and bi-objective coregionalization
matrices with unit diagonals. The coregionalization and residual
parameterizations and the positive-semidefiniteness argument are given in
Appendix~\ref{app:shared_surrogate_details}.

\paragraph{Regularized fitting.}
To control spectral complexity during sequential refitting, we fit the exact
GP to the standardized objectives $\widetilde{\mathbf Y}_n$ by minimizing
\begin{equation}
 \widehat\theta_n\in\arg\min_{\theta\in\Theta}
 \left[-\log p(\widetilde{\mathbf Y}_n\mid\mathbf X_n,\theta)
 +\mathcal R_{\mathrm{spectral}}(\theta)
 +\mathcal R_{\mathrm{other}}(\theta)\right].
 \label{eq:fit_summary}
\end{equation}
The parameter domain $\Theta$ bounds frequency locations and spectral scales.
The spectral penalty discourages rough functions, boundary-saturating
frequencies, and mixture imbalance; the remaining penalties regularize
coregionalization, residual, noise, and outputscale parameters. Each fit is
warm-started from the preceding iteration, with objective standardization
recomputed on $\mathcal D_n$. The full objective, parameter transformations,
and hyperparameter settings are specified in
Appendix~\ref{app:shared_surrogate_details}.

\paragraph{Prequential mean correction.}
We correct systematic prediction errors using residuals recorded at previously
selected candidates. Before evaluating $\mathbf x_t$, we store its raw GP
moments, the current output scale $\mathbf s_{t-1}$, and the search mode
$S_t\in\{\mathrm{local},\mathrm{global}\}$.
After observing $\mathbf y_t$, we compute
$\widetilde{\mathbf r}_t=
[\mathbf y_t-\boldsymbol\mu_{t-1}(\mathbf x_t)]\oslash\mathbf s_{t-1}$,
where $\oslash$ denotes element-wise division. For the next search mode $s$,
the correction is
\begin{equation}
 \widehat{\mathbf b}_{s,n}=
 \frac{\sum_{t\in\mathcal W_n:S_t=s}
 w_{n,t}\widetilde{\mathbf r}_t^{\,\mathrm{clip}}
 +\lambda_s\widehat{\mathbf b}_{0,n}}
 {\sum_{t\in\mathcal W_n:S_t=s}w_{n,t}+\lambda_s}.
 \label{eq:hierarchical_bias_estimate}
\end{equation}
Here $\mathcal W_n$ is a recent residual window, $w_{n,t}$ discounts older
observations, and clipping limits outlier influence. The shared estimate
$\widehat{\mathbf b}_{0,n}$ pools both modes and is shrunk toward zero;
$\lambda_s>0$ partially pools each mode toward this estimate when its own
history is limited. Every residual used for the next decision comes from a
prediction stored before the corresponding evaluation. Clipping, weighting,
and pooling are detailed in
Appendix~\ref{app:decision_posterior_details}.

\paragraph{Covariance rescaling and decision posterior.}
To address underdispersion without passing the full estimated scale correction
to acquisition optimization, we distinguish predictive and acquisition scales.
We estimate $c_{\mathrm{pred},n}$ using fixed-hyperparameter
leave-one-scalar-out and spatial cross-validation, combined with Mahalanobis
errors of past candidates' stored raw predictions. For search mode $s$, the
decision posterior is
\begin{equation}
 \begin{aligned}
 p_{\mathrm{acq}}(\mathbf f(\mathbf x)\mid\mathcal D_n,S_{n+1}=s)
 &=\mathcal N\!\left(
 \boldsymbol\mu_n(\mathbf x)+\mathbf s_n\odot\widehat{\mathbf b}_{s,n},\;
 c_{\mathrm{acq},n}\boldsymbol\Sigma_n(\mathbf x)\right),\\
 c_{\mathrm{acq},n}&=\min\{c_{\mathrm{cap}},
 c_{\mathrm{pred},n}^{\gamma}\}.
 \end{aligned}
 \label{eq:decision_posterior}
\end{equation}
Here $\odot$ denotes element-wise multiplication. Tempering and capping limit
the covariance inflation exposed to qLogEHVI. For predictive diagnostics,
we instead retain the raw mean and use
$c_{\mathrm{pred},n}\boldsymbol\Sigma_n$.
These posterior transformations leave the fitted GP unchanged, and scalar
rescaling preserves its correlation structure. Cross-validation uses
full-data hyperparameters; only the stored candidate errors are strictly
prequential. The scale estimators and their settings are given in
Appendix~\ref{app:decision_posterior_details}.

\paragraph{Pareto-local and periodic-global selection.}
We concentrate most acquisition searches around diverse observed Pareto points,
while scheduling periodic global rounds to retain full-domain exploration.
Let $\mathcal R_{n,k}(r_n)$ denote a box around the $k$-th selected Pareto
center, with a common relative half-width $r_n$. The next candidate is
obtained by
\begin{equation}
 \begin{aligned}
 \mathcal A_n&=\begin{cases}
 \displaystyle\bigcup_{k=1}^{K_n}\mathcal R_{n,k}(r_n),
 &S_{n+1}=\mathrm{local},\\
 \mathcal X,&S_{n+1}=\mathrm{global},
 \end{cases}\\
 \mathbf x_{n+1}&\approx\arg\max_{\mathbf x\in\mathcal A_n}
 \operatorname{qLogEHVI}(\mathbf x;p_{\mathrm{acq}},\mathcal P_n,
 \mathbf r_{\mathrm{HV}}),
 \end{aligned}
 \label{eq:selection_summary}
\end{equation}
where $\mathbf r_{\mathrm{HV}}$ is the acquisition reference point.
On local rounds, farthest-point selection in normalized decision and objective
coordinates determines the centers, and the restart budget is distributed
approximately equally across their boxes. We select the best numerically
optimized candidate. The common radius adapts to realized HVI on local
rounds; global rounds leave the radius and local counters unchanged and occur
more frequently early in optimization. The GP is always fitted to all
observations: only acquisition optimization is localized. Center selection,
restart allocation, scheduling, and radius updates are specified in
Appendix~\ref{app:local_global_details}.

\begin{algorithm}[t]
\caption{MOLRN-BO}
\label{alg:molrn_bo}
\begin{algorithmic}[1]
\Require Initial data $\mathcal D_{n_0}$, additional evaluation budget $T$,
         and acquisition reference point $\mathbf r_{\mathrm{HV}}$
\State Initialize empty calibration history, common local radius, and counters
\For{$n=n_0,\ldots,n_0+T-1$}
 \State Standardize objectives and fit the shared-spectral GP
        using Eqs.~\eqref{eq:composite_kernel} and~\eqref{eq:fit_summary}
 \State Set $u=n-n_0+1$; choose $S_{n+1}$ and construct $\mathcal A_n$
        using the local--global policy
        (Appendix~\ref{app:local_global_details})
 \State Estimate $\widehat{\mathbf b}_{S_{n+1},n}$ from past prequential
        residuals using Eq.~\eqref{eq:hierarchical_bias_estimate}
 \State Estimate $c_{\mathrm{pred},n}$ from cross-validation on $\mathcal D_n$
        and stored raw candidate errors
        (Appendix~\ref{app:decision_posterior_details})
 \State Construct $p_{\mathrm{acq}}$ using Eq.~\eqref{eq:decision_posterior}
 \State Obtain $\mathbf x_{n+1}$ by optimizing
        Eq.~\eqref{eq:selection_summary}
 \State Store $\boldsymbol\mu_n(\mathbf x_{n+1})$,
        $\boldsymbol\Sigma_n(\mathbf x_{n+1})$, $\mathbf s_n$, and $S_{n+1}$
 \State Evaluate $\mathbf y_{n+1}=\mathbf f(\mathbf x_{n+1})$
 \State Append the standardized residual and raw Mahalanobis error
        to the calibration history
 \State $\mathcal D_{n+1}\gets\mathcal D_n\cup
        \{(\mathbf x_{n+1},\mathbf y_{n+1})\}$
 \If{$S_{n+1}=\mathrm{local}$}
  \State Update the common radius and local counters using realized HVI
         (Appendix~\ref{app:local_global_details})
 \EndIf
\EndFor
\State \Return Nondominated observations in $\mathcal D_{n_0+T}$
\end{algorithmic}
\end{algorithm}

\section{Experiments}
\label{sec:experiments}

\subsection{Experimental settings}
\label{sec:experimental_settings}

\paragraph{Problems.}
We evaluate 12 deterministic bi-objective problems: Branin--Currin ($D=2$)
\citep{daulton2020differentiable}; WFG1, WFG2, and WFG4--WFG9 ($D=10$, $k=4$)
from the Walking Fish Group (WFG) suite~\citep{huband2006review}; and DTLZ2, DTLZ4, and DTLZ7 ($D=10$)
from the DTLZ suite~\citep{deb2002scalable}. Branin--Currin tests
sample-efficient modeling of differently shaped objective surfaces in a
low-dimensional domain; the selected WFG problems introduce combinations of
multimodality, deceptiveness, nonseparability, biased transformations, and
varied Pareto-front geometries in higher-dimensional decision spaces; and the
DTLZ problems assess front approximation under a regular concave geometry
(DTLZ2), a biased decision-to-objective mapping (DTLZ4), and a disconnected
front (DTLZ7). 

\paragraph{Comparators and budget.}
The eight baselines are independent BoTorch SingleTaskGP models
\citep{balandat2020botorch} with qLogEHVI or qLogNParEGO; lower-bound JES
\citep{tu2022joint}; MESMO \citep{belakaria2019max}; mobopt
\citep{galuzio2020mobopt}; original MO-LRN with $Q=2$ and qLogEHVI;
NSGA-II \citep{deb2002fast}; and scrambled Sobol search. To ensure statistical reliability, each algorithm–task pair was evaluated over 20 trials using random seeds 100–119.
Each problem--seed pair shares a 10-point scrambled
Sobol initial design, followed by 100 sequential evaluations (batch size one). Runs were distributed across heterogeneous
CPU/GPU systems, so no controlled runtime comparison is reported.

\paragraph{Metrics and statistical analysis.}
All methods are rescored against fixed reference fronts under a common
maximization convention, with the hypervolume reference point placed $10\%$
beyond the nadir in the worse direction along each objective. Normalized
hypervolume (nHV; higher is better) is the hypervolume of the observed
nondominated set divided by that of the fixed reference front; normalized
inverted generational distance (nIGD; lower is better) uses
ideal--nadir-normalized objectives. The four primary metrics are final nHV
and nIGD and two anytime measures: normalized trapezoidal areas under the
HV-regret and nIGD curves over sequential evaluation steps
(both AUCs are lower-is-better). Secondary analyses assess run-to-run robustness using across-seed standard
deviations and adverse-tail performance: the 10th percentile of nHV and the
90th percentiles of nIGD and both AUC measures. Additional IGD$^+$,
additive-$\epsilon$, and convergence-curve results are reported in
Appendix~\ref{sec:additional_results}.

For each primary metric, we rank algorithms by their across-seed means
within each problem and average ranks over the 12 problems. Friedman tests
assess overall algorithm differences. Comparisons between MOLRN-BO and each
baseline use Wilcoxon signed-rank tests: one-sided tests on the 12 paired
problem-level means across problems, and two-sided tests on paired
common-seed observations within each problem. Holm correction is applied
separately for each metric, across baselines for cross-problem tests and
across all problem--baseline comparisons for within-problem tests.

\subsection{Experimental results}
\label{sec:experimental_results}

\begin{table}[t]
  \centering
  \small
  \setlength{\tabcolsep}{4.2pt}
  \caption{Mean problem ranks over 12 benchmarks using all valid completed
  trials. Lower rank is better. Bold and underlined entries denote the best
  and second-best method for each metric, respectively.}
  \label{tab:overall_ranks}
  \begin{tabular}{lcccc}
    \toprule
    Method & Final nHV & Final nIGD & HV-regret AUC & nIGD AUC \\
    \midrule
    MOLRN-BO       & \textbf{2.25} & \textbf{2.17} & \underline{2.42} & \underline{2.83} \\
    qLogNParEGO    & \underline{2.42} & \underline{3.00} & \textbf{2.17} & \textbf{2.75} \\
    qLogEHVI       & 3.42 & 4.33 & 3.33 & 3.83 \\
    mobopt         & 4.67 & 4.25 & 4.92 & 4.00 \\
    JES            & 5.50 & 4.83 & 5.50 & 4.75 \\
    Original MO-LRN & 6.42 & 5.33 & 6.33 & 5.58 \\
    NSGA-II        & 5.54 & 6.50 & 6.04 & 7.00 \\
    MESMO          & 7.42 & 7.25 & 6.92 & 6.67 \\
    Sobol          & 7.38 & 7.33 & 7.38 & 7.58 \\
    \bottomrule
  \end{tabular}
\end{table}

\paragraph{Overall performance.}
As shown in Table~\ref{tab:overall_ranks}, MOLRN-BO obtains the best average
rank for both final Pareto-quality metrics: $2.25$ for nHV and $2.17$ for
nIGD. qLogNParEGO ranks second at the final budget but is slightly better on
both anytime AUC measures, placing the two methods in the leading group with
different strengths. Algorithm effects are significant for all four metrics
(Friedman $p<1.7\times10^{-6}$). 

\begin{figure}[p!]
  \centering

  \begin{subfigure}{\linewidth}
    \centering
    \includegraphics[width=\linewidth]
      {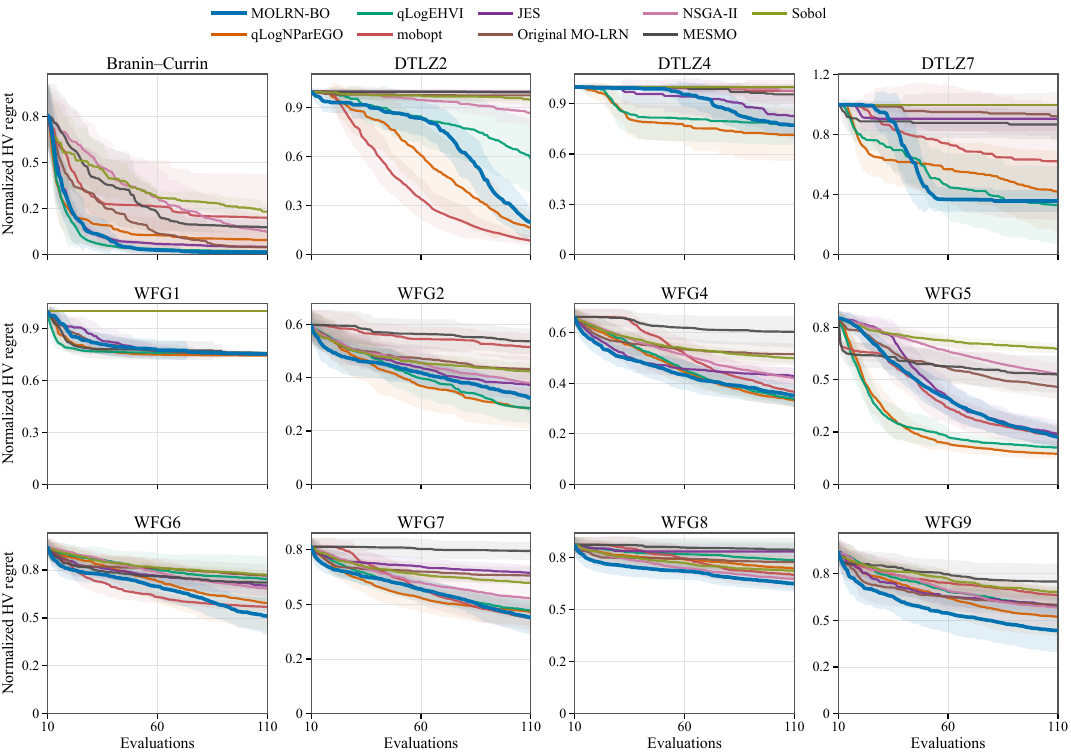}
    \caption{Normalized hypervolume regret.}
    \label{fig:nhv_regret_trajectories}
  \end{subfigure}

  \begin{subfigure}{\linewidth}
    \centering
    \includegraphics[width=\linewidth]
      {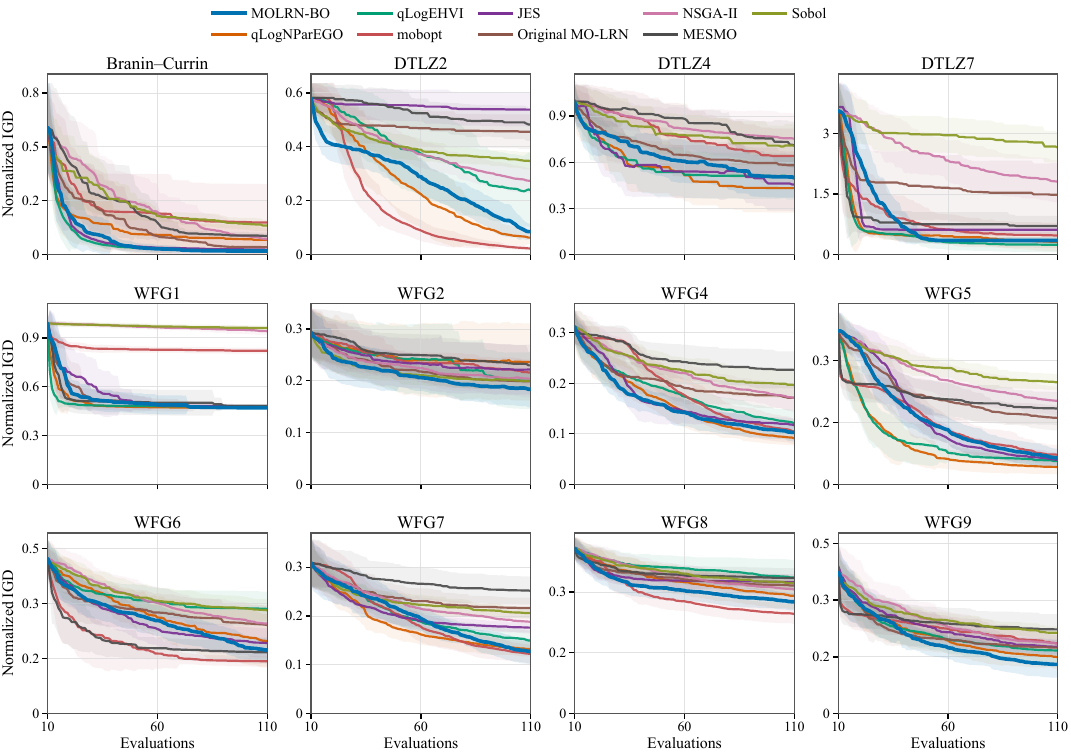}
    \caption{Normalized IGD.}
    \label{fig:nigd_trajectories}
  \end{subfigure}

  \vspace{-1mm}

  \caption{Per-problem optimization trajectories on the 12 benchmarks.
  Solid curves show across-trial means, and shaded regions indicate one
  sample standard deviation. Both metrics are lower-is-better.}
  \label{fig:performance_trajectories}
\end{figure}

Figure~\ref{fig:performance_trajectories} shows the per-problem HV-regret
and nIGD trajectories. qLogNParEGO often progresses faster initially,
whereas MOLRN-BO is competitive or leading at the final budget on many
problems. This pattern may reflect the data requirements of the spectral
surrogate and the limited early residual history. With only ten initial
observations, identifying input-dependent spectral structure, mixture
contributions, and cross-objective dependence may be less reliable than
fitting a simpler stationary surrogate. Prequential corrections also
have little residual history at this stage and are absent or strongly
shrunk. As observations accumulate, both spectral fitting and
mode-specific corrections can draw on more evidence.

\paragraph{Comparison with the original MO-LRN.}
The direct comparison with the original MO-LRN evaluates the benefit of the
complete MOLRN-BO framework. MOLRN-BO achieves better problem-level means on
$11/12$ problems for final nHV and HV-regret AUC, and on all $12/12$ problems
for final nIGD and nIGD AUC. The corresponding cross-problem tests remain
significant after Holm correction ($p\leq0.00244$). 
The contribution of each individual module is examined separately in
Sec.~\ref{sec:ablation_study}.

\paragraph{Comparison with standard MOBO baselines.}
Compared with qLogEHVI, MOLRN-BO achieves better mean final nIGD on $10/12$ problems (cross-problem $p=0.0637$), including six statistically significant improvements; all significant differences favor MOLRN-BO. It also obtains better mean final nHV on $7/12$ problems. The comparison with qLogNParEGO is closer: MOLRN-BO performs better on $6/12$ problems in final nHV and $7/12$ in final nIGD, whereas qLogNParEGO has slightly better anytime ranks. Although the aggregate differences between these two methods are not statistically significant, MOLRN-BO achieves slightly stronger final Pareto quality overall while remaining competitive throughout optimization.

\paragraph{Robustness and limitations.}
MOLRN-BO has the smallest average trial-to-trial nIGD standard deviation ($0.0358$) among the nine algorithms. It also obtains
the best adverse-tail mean rank for all four primary outcomes: $2.08$ for
nHV-q10, $2.08$ for nIGD-q90, $2.00$ for HV-regret-AUC-q90, and $2.83$ for
nIGD-AUC-q90. These results are consistent with the design objective of
reducing exposure to poor trials; module-level attribution is examined by the
ablations in Sec.~\ref{sec:ablation_study}. However, the gains are not universal: MOLRN-BO is comparatively weak on WFG5, converges more
slowly than the strongest baselines on WFG1, and exhibits slow early-to-middle
nIGD improvement on DTLZ7. Detailed robustness statistics and per-problem
curves are provided in Appendix~\ref{sec:additional_results}.

\subsection{Ablation study}
\label{sec:ablation_study}

We evaluate the principal surrogate and decision components under the experimental protocol described in Sec.~\ref{sec:experimental_settings}, using matched initial designs and the same reference fronts and normalization.

The \textbf{surrogate diagnostic} evaluates three models on a fixed set of 110
observations collected by MOLRN-BO: the original output-specific
parameterization with $Q=4$, the same parameterization with $Q=2$, and the
shared-spectrum $Q=2$ construction in Eq.~\eqref{eq:composite_kernel}.
The RBF residual is disabled in all three variants. Consequently, the first
transition isolates spectral capacity, while the second isolates spectral
sharing and per-mixture coregionalization, conditional on a common observed
design. The \textbf{decision-policy ablation} holds the shared-$Q=2$ and RBF surrogate,
covariance calibration, and qLogEHVI fixed. Its control optimizes all restarts
over the full domain and uses one bias estimate pooled across acquisition
modes. The treatment instead applies the Pareto-local/periodic-global search
policy and estimates independent biases for the local and global modes.
This comparison measures
the joint effect of search localization and mode-specific bias estimation.

Finally, the \textbf{bias-pooling ablation} holds this local--global policy
fixed and replaces the independent mode-specific estimates with the
hierarchical estimator in Eq.~\eqref{eq:hierarchical_bias_estimate}.

\begin{table}[t]
  \centering
  \small
  \setlength{\tabcolsep}{4pt}
  \caption{Controlled diagnostics and ablations over 12 benchmarks.
  Values are means over available paired problem--seed observations,
  except candidate bias, reported as the median problem-level absolute bias.
  Panel (a) disables all RBF residuals.}
  \label{tab:controlled_ablations}
  \begin{tabular}{@{}lrrrr@{}}
    \toprule
    \textbf{(a) Fixed-design surrogate}
    & \shortstack{Training\\RMSE}
    & \shortstack{Global\\RMSE}
    & \shortstack{Pareto-region\\RMSE}
    & \shortstack{Top-HVI\\optimism} \\
    \midrule
    Output-specific, $Q=4$
    & \textbf{0.018} & 1.154 & 1.101 & 1.213 \\
    Output-specific, $Q=2$
    & 0.062 & 0.816 & 0.792 & 0.498 \\
    Shared-spectrum, $Q=2$
    & 0.063 & \textbf{0.746} & \textbf{0.729} & \textbf{0.087} \\
    \midrule
    \textbf{(b) Search / bias}
    & \shortstack{Final nHV\\$\uparrow$}
    & \shortstack{Final nIGD\\$\downarrow$}
    & \shortstack{HV-regret\\AUC $\downarrow$}
    & \shortstack{nIGD\\AUC $\downarrow$} \\
    \midrule
    Full-domain / pooled
    & 0.459 & 0.255 & 0.633 & 0.349 \\
    Local--global / independent
    & \textbf{0.590} & \textbf{0.201} & \textbf{0.570} & \textbf{0.323} \\
    \midrule
    \textbf{(c) Bias pooling}
    & \shortstack{Final nHV\\$\uparrow$}
    & \shortstack{Final nIGD\\$\downarrow$}
    & \multicolumn{2}{c}{
      \shortstack{Median absolute\\candidate bias $\downarrow$}
    } \\
    \midrule
    Independent
    & \textbf{0.590} & \textbf{0.201} & \multicolumn{2}{c}{0.197} \\
    Hierarchical (MOLRN-BO)
    & 0.582 & 0.207 & \multicolumn{2}{c}{\textbf{0.079}} \\
    \bottomrule
  \end{tabular}
\end{table}

\textbf{Spectral structure.}
Reducing output-specific capacity from $Q=4$ to $Q=2$ increases training
RMSE but lowers global and Pareto-region RMSE on $10/12$ problems and
top-HVI optimism on $11/12$. At $Q=2$, the shared construction further
reduces mean global and Pareto-region RMSE and lowers mean top-HVI
optimism from $0.498$ to $0.087$, despite nearly unchanged training RMSE
(Table~\ref{tab:controlled_ablations}a). These fixed-design diagnostics
favor a lower-capacity shared surrogate for off-design prediction and
acquisition estimation.

\textbf{Decision policy.}
The local--global/independent-bias bundle improves both final and anytime
performance (Table~\ref{tab:controlled_ablations}b). Final nHV, final nIGD,
HV-regret AUC, and nIGD AUC improve on $11$, $11$, $11$, and $10$ of the
12 problems, respectively. WFG1 is the principal exception for the final
and hypervolume-based measures, and DTLZ4 for nIGD AUC.

\textbf{Bias pooling.}
Hierarchical pooling reduces candidate absolute bias on $11/12$ problems,
with its problem-level median decreasing from $0.197$ to $0.079$.
This reduction does not translate into consistent BO gains: mean final
nHV decreases from $0.590$ to $0.582$, while nIGD increases from $0.201$
to $0.207$, and neither AUC improves consistently
(Table~\ref{tab:controlled_ablations}c). The demonstrated benefit is therefore a substantial reduction in prediction bias at the cost of a slight decline in optimization performance.

\section{Conclusion}
\label{sec:conclusion}

We introduced MOLRN-BO to address the mismatch between regression fit and acquisition-driven decisions when MO-LRN is used for multi-objective optimization. For the original surrogate, lower training error can coexist with larger off-design errors and overly optimistic acquisition predictions. The framework combines a regularized shared-spectral surrogate with objective-specific residuals, prequential mean correction, tempered covariance scaling, and Pareto-local search with periodic global rounds. Across 12 deterministic bi-objective benchmarks, it significantly outperforms the original MO-LRN on all four primary metrics in Holm-adjusted cross-problem tests and achieves the best average ranks for final nHV and nIGD, with second-best anytime ranks. Fixed-design diagnostics support using a lower-capacity spectral structure shared across objectives for off-design prediction, while sequential ablations show clear gains from combining local–global search with independent bias correction. Hierarchical pooling reduces systematic candidate bias while incurring only a small loss in BO performance. Together, these findings show that adapting both surrogate structure and its use in acquisition decisions can substantially improve sequential optimization performance compared with relying on training fit alone to select a surrogate.

Despite these encouraging results, several limitations of this work should be acknowledged. The evaluation concerns small-budget, deterministic bi-objective
optimization with one candidate per round. Larger budgets and additional
outputs motivate sparse or variational inference, while noisy and batch
settings require corresponding treatment of observation uncertainty and
evaluation histories. Sensitivity to the fixed regularization settings, and the trade-off between optimization quality and computational overhead, remain to be quantified.

\subsection*{AI use statement}

In this work, we used generative AI tools for assisting with translation, cleaning and reformatting datasets, supporting qualitative and thematic data analysis or interpreting results. We have not used generative AI tools for generating synthetic data sets or conceptual frameworks, designing or providing feedback on research methodology or experiments,
and the rest of the required disclosure tasks are not applicable to this work.
Additionally, we used generative AI tools for creating and editing software code, and editing the paper to improve readability. We have reviewed all AI-assisted work. We checked that the datasets reformatted by the LLM were consistent with the original datasets and that the statistical analysis tools provided by the LLM were correct. We take responsibility for the final content of this work,
including text, claims or artifacts produced with the aid of generative AI.

\subsection*{Ethics statement}

The authors have identified no ethical concerns associated with this work.

\subsection*{Reproducibility statement}

The supplementary repository contains implementations of MOLRN-BO and all
evaluated baselines, code for reproducing the ablation studies, and
implementations of the 12 benchmark problems. Detailed formulations and
hyperparameter settings for the shared-spectral surrogate,
decision-posterior calibration, and local--global acquisition optimization
are provided in Appendix sections~\ref{app:shared_surrogate_details},
\ref{app:decision_posterior_details}, and~\ref{app:local_global_details},
respectively. Section~\ref{sec:experimental_settings} specifies the common
initial designs, random seeds, evaluation budgets, performance metrics,
and statistical tests. The ablation configurations are described in
Section~\ref{sec:ablation_study}, and additional performance and robustness
analyses are reported in Appendix~\ref{sec:additional_results}.

\bibliography{iclr2027_conference}
\bibliographystyle{iclr2027_conference}

\appendix

\section{Preliminary diagnoses}
\label{app:pilot_diagnostics}

The original parameterization of MO-LRN assigns separate spectral
factors to each output, with cross-output covariance induced by their spectral overlap. However, pilot sequential experiments revealed five BO-oriented problems when MO-LRN
was used directly in MOBO. Three arose at the surrogate level: in-sample fit
was poorly aligned with candidate-region accuracy, learned frequencies
frequently approached their imposed bounds, and cross-output spectral overlap
could become numerically negligible. A fourth concerned the decision
posterior, which exhibited biased and underdispersed predictions at
acquisition-selected points. The fifth arose during acquisition optimization,
where unrestricted full-domain search favored distant and boundary-adjacent
candidates. Table~\ref{tab:pilot_diagnostics} summarizes these diagnostics on
the ten-dimensional WFG4 development problem.

These failure modes together motivate three corresponding changes in Section~\ref{sec:method}: a
more structured and regularized surrogate, an error-calibrated posterior used only for decisions, and an acquisition search strategy that distributes local refinement across the Pareto set while retaining global exploration.

\begin{table*}[th!]
    \centering
    \caption{
        Preliminary BO-oriented diagnostics on the ten-dimensional WFG4
        development problem. These descriptive pilot results motivate the
        design of MOLRN-BO; broader controlled evaluations are presented in
        Section~4.3.
    }
    \label{tab:pilot_diagnostics}
    \small
    \setlength{\tabcolsep}{5pt}
    \renewcommand{\arraystretch}{1.15}
    \begin{tabularx}{\textwidth}{
        >{\raggedright\arraybackslash}p{0.19\textwidth}
        >{\raggedright\arraybackslash}p{0.29\textwidth}
        >{\raggedright\arraybackslash}X
    }
        \toprule
        Failure mode
        & Pilot experiment and sample
        & Key result \\
        \midrule

        Capacity--decision mismatch
        &
        WFG4 at \(n=110\); three data trajectories and five model fits
        per trajectory; \(Q=4\) and \(Q=2\) fitted to identical data
        &
        \(Q=4\) vs.\ \(Q=2\):
        train RMSE \(0.0022\) vs.\ \(0.0082\);
        global RMSE \(0.9518\) vs.\ \(0.7730\);
        Pareto-region RMSE \(0.9372\) vs.\ \(0.8530\).
        \\
        \midrule
        Boundary-saturating spectra
        &
        Original \(Q=4\) MO-LRN on WFG4;
        300 sequential fits with \(f_{\max}=8\)
        &
        At least one learned frequency was within \(5\%\) of the imposed
        bound in \(296/300\) fits (\(98.7\%\)); the mean per-fit maximum
        frequency was \(7.96\), and normalized roughness averaged \(59.6\).
        \\
        \midrule
        Vanishing spectral overlap
        &
        Original \(Q=4\) MO-LRN on ten-dimensional WFG4;
        spectral snapshots at \(n=20,50,80,\) and \(110\)
        &
        Fractions of mixtures with
        \(R_{q,12}<10^{-12}\) were
        \(47\%, 95\%, 100\%,\) and \(100\%\), respectively.
        \\
        \midrule
        Biased and overconfident predictions
        &
        Original \(Q=4\) MO-LRN on WFG4;
        300 strictly prequential predictions at acquisition-selected points
        &
        Normalized RMSE was \(3.90\), while the empirical coverage of
        nominal \(90\%\) posterior intervals was only \(3.7\%\).
        \\
        \midrule
        Extrapolative acquisition search
        &
        WFG4; three paired pilot trials and 300 acquisition decisions
        per policy
        &
        Full-domain vs.\ local--global search:
        nearest-data distance \(0.741\) vs.\ \(0.497\);
        boundary-coordinate fraction \(24.5\%\) vs.\ \(16.7\%\).
        \\

        \bottomrule
    \end{tabularx}

    \vspace{2pt}
    \begin{minipage}{\textwidth}
        \footnotesize
        \textit{Notes.}
        The normalized spectral overlap \(R_{q,12}\in[0,1]\) measures
        the overlap between the two output-specific spectral factors of
        mixture \(q\). Model fits in the first row share the same data.
    \end{minipage}
\end{table*}

\section{Calculation Detail of MOLRN-BO}

We consider the maximization of an expensive vector-valued black-box function
$\mathbf f:\mathcal X\subset\mathbb R^D\rightarrow\mathbb R^M$. After $n$
evaluations, the available data are
$\mathcal D_n=\{(\mathbf x_t,\mathbf y_t)\}_{t=1}^n$, where
$\mathbf y_t=\mathbf f(\mathbf x_t)+\boldsymbol\epsilon_t$. 

\subsection{Regularized shared-spectral surrogate}
\label{app:shared_surrogate_details}

The first three diagnostics in Appendix~\ref{app:pilot_diagnostics} suggest that
using MO-LRN reliably in sequential optimization requires controlling its
spectral capacity while preventing cross-objective transfer from depending on
the overlap of independently fitted output spectra. We therefore separate
nonstationary input-space structure, cross-objective dependence, and
objective-specific smooth variation within a single composite surrogate.

At BO iteration \(n\), each objective is standardized using the observations
available before selecting the next candidate:
\begin{equation}
  \bar y_{n,i}
  =
  \frac{1}{n}\sum_{j=1}^{n}y_{j,i},
  \qquad
  s_{n,i}
  =
  \max\!\left\{
    \left[
      \frac{1}{n}\sum_{j=1}^{n}
      (y_{j,i}-\bar y_{n,i})^2
    \right]^{1/2},
    \varepsilon_Y
  \right\},
  \qquad
  \widetilde y_{j,i}
  =
  \frac{y_{j,i}-\bar y_{n,i}}{s_{n,i}},
  \label{eq:output_standardization}
\end{equation}
where \(\varepsilon_Y>0\) prevents division by zero. The GP is fitted to
\(\widetilde{\mathbf Y}_n\), while its posterior moments are transformed back
to the original objective scale before acquisition construction.

The spectral locations, scales, mixture weights, coregionalization,
and residual parameters are learned by regularized marginal-likelihood
optimization. Parameter bounds, penalty strengths, and the calibration
and search settings specified below are fixed hypoerparameter configurations.

\paragraph{Composite covariance.}
Equation~\eqref{eq:composite_kernel} combines shared nonstationary input
covariance, a common cross-objective correlation matrix for all mixtures,
and independent smooth residuals. The shared scalar components follow
the real four-term MO-LRN construction~\citep{xu2026revisiting}, with
spectral-factor parameters shared across outputs. For each mixture $q$
and coordinate $d$, let $\mu_{q,d}^{(a)}$ and
$\sigma_{q,d}^{(a)}>0$, $a\in\{1,2\}$, denote the Gaussian
spectral-factor means and standard deviations. The index $a$
distinguishes the two frequency arguments, not the outputs:
the parameters are shared across objectives, but those for $a=1$
and $a=2$ need not coincide. Frequency pairs are independent across
coordinates, with a within-pair correlation $\zeta_q$ shared across
coordinates within each mixture. Define
\begin{equation}
 C_{q,d}
 =
 \begin{bmatrix}
  (\sigma_{q,d}^{(1)})^2
  &
  \zeta_q\sigma_{q,d}^{(1)}\sigma_{q,d}^{(2)}
  \\
  \zeta_q\sigma_{q,d}^{(1)}\sigma_{q,d}^{(2)}
  &
  (\sigma_{q,d}^{(2)})^2
 \end{bmatrix},
 \qquad
 V_{q,d}^{(0)}=\frac12 C_{q,d}.
 \label{eq:shared_spectral_covariance}
\end{equation}
The spectral correlation is parameterized as
\[
 \zeta_q
 =
 (1-\epsilon_\zeta)
 \left[
  2\operatorname{sigmoid}(\widetilde\zeta_q)-1
 \right],
 \qquad
 \epsilon_\zeta=10^{-6},
\]
which ensures $|\zeta_q|<1$ and $C_{q,d}\succ0$.
It is distinct from the coregionalization parameter $r$ and the relative
residual scale $\rho_i$ defined below.

Let
$\boldsymbol\mu_{q,d}
=(\mu_{q,d}^{(1)},\mu_{q,d}^{(2)})^{\mathsf T}$.
The normalized product of the two identical Gaussian spectral factors
has mean $\boldsymbol\mu_{q,d}$ and covariance
$V_{q,d}^{(0)}=C_{q,d}/2$. For numerical stability, determinants in
the inverse calculations are bounded below by $\tau=10^{-8}$.
Writing $\operatorname{adj}$ for the $2\times2$ adjugate, we define
the effective spectral moments by
\begin{align}
 P_{q,d}
 &=
 \frac{\operatorname{adj}(C_{q,d})}
      {\max\{\det C_{q,d},\tau\}},
 \nonumber\\
 T_{q,d}
 &=2P_{q,d},
 \qquad
 V_{q,d}
 =
 \frac{\operatorname{adj}(T_{q,d})}
      {\max\{\det T_{q,d},\tau\}},
 \nonumber\\
 \boldsymbol\nu_{q,d}
 &=
 V_{q,d}(2P_{q,d})\boldsymbol\mu_{q,d}.
 \label{eq:shared_effective_moments}
\end{align}
The determinant floor is applied separately to each factor covariance
and to the summed precision. Under spectral sharing, the two
stabilized factor precisions are identical, yielding $T_{q,d}=2P_{q,d}$.
If neither floor is active, these expressions reduce to
$V_{q,d}=C_{q,d}/2$ and
$\boldsymbol\nu_{q,d}=\boldsymbol\mu_{q,d}$.
The parameters $\sigma_{q,d}^{(a)}$ describe the factors before
multiplication; the fourth-moment regularization in
Eq.~\eqref{eq:roughness_penalty} acts on these pre-product parameters.

Write
\[
 V_q^{ab}
 =
 \operatorname{diag}_{d=1,\ldots,D}
 \!\left([V_{q,d}]_{ab}\right),
 \qquad
 \boldsymbol\nu_q^{(a)}
 =
 \left(\nu_{q,d}^{(a)}\right)_{d=1}^{D},
 \qquad
 \boldsymbol\delta=\mathbf x-\mathbf x'.
\]
Let $A>0$ be the shared-kernel outputscale and let $\pi_q>0$ be
the normalized mixture weights in Eq.~\eqref{eq:mixture_weights},
with $\sum_q\pi_q=1$. The shared scalar component is
\begin{align}
 F_q(\mathbf x,\mathbf x')
 &=
 \exp\!\left[
  -\frac12\left(
   \mathbf x^{\mathsf T}V_q^{11}\mathbf x
   -2\mathbf x^{\mathsf T}V_q^{12}\mathbf x'
   +\mathbf x'^{\mathsf T}V_q^{22}\mathbf x'
  \right)
 \right]
 \nonumber\\
 &\qquad\times
 \cos\!\left(
  (\boldsymbol\nu_q^{(1)})^{\mathsf T}\mathbf x
  -(\boldsymbol\nu_q^{(2)})^{\mathsf T}\mathbf x'
 \right),
 \nonumber\\
 G_q^{(a)}(\boldsymbol\delta)
 &=
 \exp\!\left[
  -\frac12\boldsymbol\delta^{\mathsf T}
  V_q^{aa}\boldsymbol\delta
 \right]
 \cos\!\left(
  (\boldsymbol\nu_q^{(a)})^{\mathsf T}\boldsymbol\delta
 \right),
 \qquad a\in\{1,2\},
 \nonumber\\
 k_q^{\mathrm{sLRN}}(\mathbf x,\mathbf x')
 &=
 \frac{A\pi_qg_q}{4}
 \left[
  F_q(\mathbf x,\mathbf x')
  +F_q(\mathbf x',\mathbf x)
  +G_q^{(1)}(\boldsymbol\delta)
  +G_q^{(2)}(\boldsymbol\delta)
 \right],
 \label{eq:shared_four_term}
\end{align}
where $g_q>0$ is the amplitude coefficient defined below.
The frequency parameters are angular frequencies, so no additional
$2\pi$ factor appears in the phase terms.

With determinant floors inactive and in the zero-overlap-jitter limit,
the amplitude coefficient is
\[
 g_q^{(0)}
 =
 (2\pi)^{-D/2}
 \prod_{d=1}^{D}\det(2C_{q,d})^{-1/2}.
\]
This coefficient is $(2\pi)^{D/2}$ times the overlap integral of two
identical, unit-normalized $2D$-variate Gaussian spectral factors.
For the stabilized amplitude, we use $\epsilon_o=10^{-6}$ and the same
determinant floor $\tau=10^{-8}$:
\begin{align}
 S_{q,d}
 &=
 2C_{q,d}+\epsilon_o I_2,
 \nonumber\\
 M_{q,d}
 &=
 \frac{\operatorname{adj}(S_{q,d})}
      {\max\{\det S_{q,d},\tau\}},
 \nonumber\\
 g_q
 &=
 (2\pi)^{-D/2}
 \prod_{d=1}^{D}
 \left[
  \max\{
   \det(M_{q,d}+\epsilon_o I_2),\tau
  \}
 \right]^{1/2}.
 \label{eq:shared_amplitude}
\end{align}
The overlap jitter regularizes both the covariance sum and its
stabilized inverse, with determinant floors applied at both stages.
These operations affect only $g_q$; the effective moments
$V_{q,d}$ and $\boldsymbol\nu_{q,d}$ remain those defined in
Eq.~\eqref{eq:shared_effective_moments}, without overlap jitter.
The determinant floors in the effective-moment calculation can,
however, alter the covariance and mean entering the exponential and
phase terms in Eq.~\eqref{eq:shared_four_term}.
When the amplitude determinant floors are inactive,
Eq.~\eqref{eq:shared_amplitude} becomes
\[
 g_q
 =
 (2\pi)^{-D/2}
 \prod_{d=1}^{D}
 \det\!\left[
  (2C_{q,d}+\epsilon_o I_2)^{-1}
  +\epsilon_o I_2
 \right]^{1/2},
\]
which converges to $g_q^{(0)}$ as $\epsilon_o\to0$, provided the
floors remain inactive.

For the bi-objective problems considered here, all mixtures use the
same unit-diagonal coregionalization matrix
\begin{equation}
 \mathbf B
 =
 \begin{bmatrix}
  1&r\\
  r&1
 \end{bmatrix},
 \qquad
 r=0.995\tanh(\widetilde r).
 \label{eq:coregionalization}
\end{equation}
Since $|r|<1$, $\mathbf B$ is positive definite.
With $\alpha_q=\sqrt{A\pi_q}$, the spectral-factor product contributes
$A\pi_q$ once through Eq.~\eqref{eq:shared_four_term}. The unit-diagonal
matrix $\mathbf B$ controls cross-objective correlation. Each mixture's
variance contribution also depends on $g_q$; at the origin,
$k_q^{\mathrm{sLRN}}(\mathbf 0,\mathbf 0)=A\pi_qg_q$.

The positive semidefiniteness of the scalar component follows from
a feature representation. The stabilized covariance $V_{q,d}$ is
positive definite: $C_{q,d}\succ0$, and both adjugate divisions in
Eq.~\eqref{eq:shared_effective_moments} have positive denominators.
Let the frequency pairs be distributed independently across
coordinates as
\[
 \begin{pmatrix}
  \omega_{q,d}^{(1)}\\
  \omega_{q,d}^{(2)}
 \end{pmatrix}
 \sim
 \mathcal N_2\!\left(
  \boldsymbol\nu_{q,d},
  V_{q,d}
 \right),
\]
and write
$\boldsymbol\omega_q^{(a)}
=(\omega_{q,d}^{(a)})_{d=1}^{D}$.
Define
\[
 \boldsymbol\phi_q(\mathbf x)
 =
 \frac12
 \begin{pmatrix}
  \cos((\boldsymbol\omega_q^{(1)})^{\mathsf T}\mathbf x)
  +\cos((\boldsymbol\omega_q^{(2)})^{\mathsf T}\mathbf x)
  \\
  \sin((\boldsymbol\omega_q^{(1)})^{\mathsf T}\mathbf x)
  +\sin((\boldsymbol\omega_q^{(2)})^{\mathsf T}\mathbf x)
 \end{pmatrix}.
\]
Expanding the expected inner product gives
\begin{equation}
 k_q^{\mathrm{sLRN}}(\mathbf x,\mathbf x')
 =
 A\pi_qg_q\,
 \mathbb E\!\left[
  \boldsymbol\phi_q(\mathbf x)^{\mathsf T}
  \boldsymbol\phi_q(\mathbf x')
 \right].
 \label{eq:shared_feature_representation}
\end{equation}
Hence, $k_q^{\mathrm{sLRN}}$ is positive semidefinite for every
admissible set of spectral parameters and every positive scalar
amplitude $g_q$.

The residual kernel for objective \(i\) is an ARD-RBF covariance,
\begin{equation}
  k_i^{\mathrm{RBF}}(\mathbf x,\mathbf x')
  =
  a_i
  \exp\!\left[
    -\frac{1}{2}
    \sum_{d=1}^{D}
    \frac{(x_d-x_d')^2}{\ell_{i,d}^2}
  \right],
  \qquad
  a_i>0,\quad \ell_{i,d}>0.
  \label{eq:rbf_residual}
\end{equation}
In the implementation, \(a_i=\rho_i v_i^{\mathrm{shared}}\), where
\(v_i^{\mathrm{shared}}\) is a reference marginal variance of the shared
kernel and \(\rho_i\) is a learned relative residual scale. This prevents the
RBF component from becoming numerically negligible merely because the
marginal scale of the LRN kernel changes with the input dimension.

For any finite input set, the contribution of component \(q\) has Gram matrix
\(\mathbf B_q\otimes\mathbf K_q^{\mathrm{sLRN}}\), which is positive
semidefinite because both factors are positive semidefinite. The residual
kernels form a block-diagonal positive-semidefinite matrix. Equation
\eqref{eq:composite_kernel}, being a sum of these terms, therefore defines a
valid matrix-valued covariance kernel.

\paragraph{Spectral constraints and regularization.}
To prevent unstable high-frequency fits, the frequency locations and spectral
scales are represented by bounded transformations:
\begin{align}
  \mu_{q,d}^{(a)}
  &=
  \mu_{\max}
  \tanh\!\left(\widetilde\mu_{q,d}^{(a)}\right),
  \qquad a\in\{1,2\},
  \label{eq:bounded_frequency}\\
  \sigma_{q,d}^{(a)}
  &=
  \sigma_{\min}
  +
  (\sigma_{\max}-\sigma_{\min})
  \operatorname{sigmoid}
  \!\left(\widetilde\sigma_{q,d}^{(a)}\right).
  \label{eq:bounded_spectral_scale}
\end{align}
The normalized mixture weights are
\begin{equation}
  \pi_q
  =
  \pi_{\min}
  +
  (1-Q\pi_{\min})
  \frac{\exp(\widetilde\pi_q)}
       {\sum_{q'=1}^{Q}\exp(\widetilde\pi_{q'})},
  \qquad
  \sum_{q=1}^{Q}\pi_q=1.
  \label{eq:mixture_weights}
\end{equation}
The lower bound \(\pi_{\min}\) prevents a component from disappearing
completely during optimization.

For a Gaussian spectral coordinate, define its fourth moment as
\begin{equation}
  m_4(\mu,\sigma)
  =
  \mu^4+6\mu^2\sigma^2+3\sigma^4.
  \label{eq:fourth_spectral_moment}
\end{equation}
We penalize the aggregate normalized fourth moment,
\begin{equation}
  \mathcal R_{\mathrm{rough}}
  =
  \frac{\lambda_{\mathrm{rough}}}{L_{\mathrm{spec}}^4}
  \sum_{q=1}^{Q}\sum_{d=1}^{D}
  \sum_{a=1}^{2}
  m_4\!\left(
    \mu_{q,d}^{(a)},\sigma_{q,d}^{(a)}
  \right),
  \qquad
  L_{\mathrm{spec}}
  =
  \max\{\mu_{\max},\sigma_{\max}\}.
  \label{eq:roughness_penalty}
\end{equation}
This discourages spectral solutions that imply unnecessarily rough
input-space functions.

In addition to the bounded transformation in
Eq.~\eqref{eq:bounded_frequency}, we penalize frequencies near the limits
of their allowed range:
\begin{equation}
  \mathcal R_{\mathrm{bdry}}
  =
  \frac{\lambda_{\mathrm{bdry}}}{2}
  \sum_{a=1}^{2}
  \operatorname{mean}_{q,d}
  \left[
    -\log
    \max\!\left\{
      \varepsilon_{\mathrm{bdry}},
      1-
      \left(
        \frac{\mu_{q,d}^{(a)}}{\mu_{\max}}
      \right)^2
    \right\}
  \right].
  \label{eq:frequency_barrier}
\end{equation}
Mixture collapse is additionally discouraged by
\begin{equation}
  \mathcal R_{\mathrm{mix}}
  =
  -\lambda_{\mathrm{mix}}
  \sum_{q=1}^{Q}\log(Q\pi_q),
  \label{eq:mixture_balance}
\end{equation}
which is minimized at uniform mixture weights.

The unconstrained coregionalization parameters receive a zero-centered
Gaussian penalty,
\begin{equation}
  \mathcal R_B
  =
  \frac{1}{2\sigma_B^2}
  \sum_{q=1}^{Q}\widetilde r_q^2.
  \label{eq:coregularization_penalty}
\end{equation}
For positive parameters, we use the generic log-scale penalty
\begin{equation}
  \mathcal P(\mathbf z;\mathbf z_0,\sigma_z)
  =
  \frac{1}{2}
  \left\|
    \frac{\log\mathbf z-\log\mathbf z_0}{\sigma_z}
  \right\|_2^2.
  \label{eq:log_scale_penalty}
\end{equation}
The residual, observation-noise, and shared-outputscale penalties are then
\begin{align}
  \mathcal R_{\mathrm{RBF}}
  &=
  \mathcal P(\boldsymbol\rho;
             \rho_0\mathbf 1,\sigma_\rho)
  +
  \mathcal P(\boldsymbol\ell;
             \ell_0\mathbf 1,\sigma_\ell),
  \label{eq:rbf_regularization}\\
  \mathcal R_{\mathrm{noise}}
  &=
  \mathcal P(\boldsymbol\eta;
             \eta_0\mathbf 1,\sigma_\eta),
  \qquad
  \mathcal R_{\mathrm{out}}
  =
  \mathcal P(A;A_0,\sigma_A),
  \label{eq:scale_regularization}
\end{align}
where \(\boldsymbol\eta\) denotes the objective-wise noise levels and \(A\)
is the shared-kernel output scale.

The composite exact GP is fitted by minimizing
\begin{equation}
  \mathcal J_n(\theta)
  =
  -\log p(
    \widetilde{\mathbf Y}_n
    \mid
    \mathbf X_n,\theta)
  +
  \mathcal R_{\mathrm{rough}}
  +
  \mathcal R_{\mathrm{bdry}}
  +
  \mathcal R_{\mathrm{mix}}
  +
  \mathcal R_B
  +
  \mathcal R_{\mathrm{RBF}}
  +
  \mathcal R_{\mathrm{noise}}
  +
  \mathcal R_{\mathrm{out}}.
  \label{eq:map_objective}
\end{equation}
The solution from the preceding BO iteration is used to initialize the next
fit, while the objective standardization in
Eq.~\eqref{eq:output_standardization} is recomputed from the currently
available observations.

In all experiments, \(Q=2\), \(\mu_{\max}=4\),
\((\sigma_{\min},\sigma_{\max})=(0.02,8)\),
\(\pi_{\min}=0.05\), \(\lambda_{\mathrm{rough}}=1\),
\(\lambda_{\mathrm{mix}}=1\),
\(\lambda_{\mathrm{bdry}}=0.5\),
\(\varepsilon_{\mathrm{bdry}}=10^{-6}\), and \(\sigma_B=1.5\).
The RBF relative scale is initialized at \(\rho_0=0.05\), with bounds
\([10^{-5},10]\) and log-prior width \(\sigma_\rho=0.75\).
Its ARD lengthscales are initialized at \(\ell_0=0.5\), with bounds
\([0.02,10]\) and log-prior width \(\sigma_\ell=1.5\).

\subsection{Mean correction and covariance-scale adjustment}
\label{app:decision_posterior_details}

The surrogate modifications above regularize the fitted covariance structure,
but they do not ensure accurate uncertainty at the points preferred by the
acquisition function. The fourth diagnostic in Appendix~\ref{app:pilot_diagnostics} shows
systematic mean error and posterior underdispersion at
acquisition-selected points. We address these errors through mean
correction and covariance rescaling of the fitted GP posterior
for acquisition decisions.

Suppose the candidate selected at BO iteration \(t\) is \(\mathbf x_t\).
Before evaluating it, we record the raw GP predictive moments
\begin{equation}
  \boldsymbol\mu_t
  =
  \boldsymbol\mu_{\mathrm{GP},t-1}(\mathbf x_t),
  \qquad
  \boldsymbol\Sigma_t
  =
  \boldsymbol\Sigma_{\mathrm{GP},t-1}(\mathbf x_t),
  \label{eq:stored_prequential_prediction}
\end{equation}
together with the current output scale \(\mathbf s_{t-1}\) and search mode
\(S_t\). Only after observing \(\mathbf y_t\) do we append the resulting error
to the calibration history. Thus, every correction used to select
\(\mathbf x_t\) depends only on errors observed before iteration \(t\).

\paragraph{Hierarchical mean-bias correction.}
The standardized prequential residual is
\begin{equation}
  \widetilde{\mathbf r}_t
  =
  (\mathbf y_t-\boldsymbol\mu_t)
  \oslash\mathbf s_{t-1},
  \label{eq:standardized_prequential_residual}
\end{equation}
where \(\oslash\) denotes element-wise division. Let \(\mathcal W_n\) contain
the most recent \(W_b\) residuals available before iteration \(n+1\).
For each objective, we compute the component-wise median and MAD,
\begin{equation}
  \mathbf m_n
  =
  \operatorname{median}_{t\in\mathcal W_n}
  \widetilde{\mathbf r}_t,
  \qquad
  \mathbf d_n
  =
  \operatorname{median}_{t\in\mathcal W_n}
  \left|
    \widetilde{\mathbf r}_t-\mathbf m_n
  \right|,
  \qquad
  \boldsymbol\tau_n
  =
  \max\{1.4826\mathbf d_n,\tau_{\min}\mathbf 1\}.
  \label{eq:residual_robust_scale}
\end{equation}
The residuals are clipped component-wise according to
\begin{equation}
  \widetilde{\mathbf r}_t^{\,\mathrm{clip}}
  =
  \operatorname{clip}\!\left(
    \widetilde{\mathbf r}_t,\,
    \mathbf m_n-\kappa_b\boldsymbol\tau_n,\,
    \mathbf m_n+\kappa_b\boldsymbol\tau_n
  \right).
  \label{eq:clipped_prequential_residual}
\end{equation}
Recent observations receive exponentially decaying weights
\begin{equation}
  w_{n,t}
  =
  2^{-a_{n,t}/H_b},
  \label{eq:bias_temporal_weight}
\end{equation}
where \(a_{n,t}=0\) for the newest residual in \(\mathcal W_n\), \(1\) for
the second newest, and so forth, and \(H_b\) is the half-life.

A shared bias estimate is first computed from all search modes:
\begin{equation}
  \widehat{\mathbf b}_{0,n}
  =
  \frac{
    \sum_{t\in\mathcal W_n}
    w_{n,t}
    \widetilde{\mathbf r}_t^{\,\mathrm{clip}}
  }{
    \lambda_0+
    \sum_{t\in\mathcal W_n}w_{n,t}
  }.
  \label{eq:shared_bias_estimate}
\end{equation}
For the mode \(s\in\{\mathrm{local},\mathrm{global}\}\) planned for the next
iteration, the mode-specific estimate is partially pooled toward the shared
estimate:
\begin{equation}
  \widehat{\mathbf b}_{s,n}
  =
  \frac{
    \displaystyle
    \sum_{\substack{t\in\mathcal W_n\\S_t=s}}
    w_{n,t}
    \widetilde{\mathbf r}_t^{\,\mathrm{clip}}
    +
    \lambda_s\widehat{\mathbf b}_{0,n}
  }{
    \displaystyle
    \sum_{\substack{t\in\mathcal W_n\\S_t=s}}
    w_{n,t}
    +
    \lambda_s
  }.
  \label{eq:app_hierarchical_bias_estimate}
\end{equation}
When few global evaluations are available, this estimate remains close to the
shared residual trend. As mode-specific evidence accumulates, the corresponding
local or global correction receives greater weight.

\paragraph{Fixed-hyperparameter spatial cross-validation.}
We next estimate covariance-scale error under spatial extrapolation. Let
\(\mathbf K_n\) be the predictive training covariance, including observation
noise, evaluated at the fitted hyperparameters, and let
\begin{equation}
  \mathbf P_n=\mathbf K_n^{-1},
  \qquad
  \boldsymbol\alpha_n
  =
  \mathbf P_n
  \operatorname{vec}(\widetilde{\mathbf Y}_n).
  \label{eq:training_precision}
\end{equation}
The leave-one-scalar-out Mahalanobis sum can be obtained directly from the
precision matrix:
\begin{equation}
  H_{\mathrm{LOO}}
  =
  \sum_{a=1}^{nM}
  \frac{\alpha_{n,a}^2}{P_{n,aa}}.
  \label{eq:loo_mahalanobis}
\end{equation}

To form spatial folds, we center the observed inputs and compute their first
principal direction,
\begin{equation}
  \mathbf v_1
  \in
  \arg\max_{\|\mathbf v\|_2=1}
  \mathbf v^{\mathsf T}
  \mathbf X_{n,c}^{\mathsf T}
  \mathbf X_{n,c}
  \mathbf v.
  \label{eq:spatial_cv_direction}
\end{equation}
The observations are sorted by their projection
\(\mathbf X_{n,c}\mathbf v_1\) and divided into \(G\) contiguous folds.
Let \(\mathcal G_g\) contain the stacked scalar-output indices associated
with fold \(g\). With the GP hyperparameters fixed at their full-data values,
the conditional residual and covariance for this fold are
\begin{equation}
  \mathbf e_g
  =
  \mathbf P_{n,\mathcal G_g\mathcal G_g}^{-1}
  \boldsymbol\alpha_{n,\mathcal G_g},
  \qquad
  \mathbf V_g
  =
  \mathbf P_{n,\mathcal G_g\mathcal G_g}^{-1}.
  \label{eq:fixed_hyperparameter_fold}
\end{equation}
Consequently,
\begin{equation}
  H_{\mathrm{sp}}
  =
  \sum_{g=1}^{G}
  \mathbf e_g^{\mathsf T}
  \mathbf V_g^{-1}
  \mathbf e_g
  =
  \sum_{g=1}^{G}
  \boldsymbol\alpha_{n,\mathcal G_g}^{\mathsf T}
  \mathbf P_{n,\mathcal G_g\mathcal G_g}^{-1}
  \boldsymbol\alpha_{n,\mathcal G_g}.
  \label{eq:spatial_cv_mahalanobis}
\end{equation}
No hyperparameters are refitted for individual folds. Because the fitted hyperparameters depend on the held-out observations,
these residuals provide internal scale diagnostics rather than an
independent assessment of predictive coverage.

For a Mahalanobis sum \(H\) containing \(N_H\) scalar residuals, define the
shrunk covariance-scale estimate
\begin{equation}
  c(H,N_H)
  =
  \operatorname{clip}_{[c_{\min},c_{\max}^{\mathrm{CV}}]}
  \left(
    \frac{\lambda_{\mathrm{CV}}c_0+H}
         {\lambda_{\mathrm{CV}}+N_H}
  \right).
  \label{eq:shrunk_cv_scale}
\end{equation}
The LOO and spatial estimates are combined geometrically:
\begin{equation}
  c_{\mathrm{LOO}}
  =
  c(H_{\mathrm{LOO}},nM),
  \qquad
  c_{\mathrm{sp}}
  =
  c(H_{\mathrm{sp}},nM),
  \qquad
  c_{\mathrm{CV}}
  =
  c_{\mathrm{LOO}}^{\,1-\omega}
  c_{\mathrm{sp}}^{\,\omega}.
  \label{eq:combined_cv_scale}
\end{equation}

\paragraph{Acquisition-selected uncertainty error.}
For each previously evaluated acquisition-selected point, we calculate the raw
normalized Mahalanobis error
\begin{equation}
  h_t
  =
  \frac{1}{M}
  (\mathbf y_t-\boldsymbol\mu_t)^{\mathsf T}
  \boldsymbol\Sigma_t^{-1}
  (\mathbf y_t-\boldsymbol\mu_t).
  \label{eq:mahalanobis_error}
\end{equation}
Under a correctly scaled \(M\)-variate predictive distribution,
\(\mathbb E[h_t]\) is approximately one.

Let \(\mathcal U_n\) contain the most recent \(W_c\) valid values. We first
define
\begin{equation}
  u_t
  =
  \log
  \operatorname{clip}_{[1,c_{\max}^{\mathrm{sel}}]}(h_t),
  \qquad
  u_{\mathrm{cap}}
  =
  \min\!\left\{
    \log c_{\max}^{\mathrm{sel}},
    \operatorname{median}_{t\in\mathcal U_n}u_t
    +
    1.4826\kappa_c
    \operatorname{MAD}_{t\in\mathcal U_n}(u_t)
  \right\},
  \label{eq:robust_log_mahalanobis}
\end{equation}
and winsorize only the upper tail,
\begin{equation}
  u_t^{\mathrm{clip}}
  =
  \min\{u_t,u_{\mathrm{cap}}\}.
  \label{eq:winsorized_log_mahalanobis}
\end{equation}
Using weights \(v_{n,t}=2^{-a_{n,t}/H_c}\), the selected-point scale is
\begin{equation}
  c_{\mathrm{sel}}
  =
  \operatorname{clip}_{[1,c_{\max}^{\mathrm{sel}}]}
  \left[
    \exp\!\left(
      \frac{
        \sum_{t\in\mathcal U_n}
        v_{n,t}u_t^{\mathrm{clip}}
      }{
        \lambda_c+
        \sum_{t\in\mathcal U_n}v_{n,t}
      }
    \right)
  \right].
  \label{eq:selected_point_scale}
\end{equation}
The full predictive covariance scale is
\begin{equation}
  c_{\mathrm{pred}}
  =
  \min\!\left\{
    c_{\max}^{\mathrm{sel}},
    \max\{1,c_{\mathrm{CV}},c_{\mathrm{sel}}\}
  \right\}.
  \label{eq:predictive_scale}
\end{equation}

Using \(c_{\mathrm{pred}}\) directly in qLogEHVI can overemphasize uncertain
regions. The covariance scale exposed to the acquisition function is therefore
tempered and capped:
\begin{equation}
  c_{\mathrm{acq}}
  =
  \min\!\left\{
    c_{\mathrm{cap}},
    c_{\mathrm{pred}}^\gamma
  \right\}.
  \label{eq:acquisition_scale}
\end{equation}
This gives two posterior views:
\begin{align}
  p_{\mathrm{pred}}(\mathbf f(\mathbf x)\mid\mathcal D_n)
  &=
  \mathcal N\!\left(
    \boldsymbol\mu_{\mathrm{GP}}(\mathbf x),
    c_{\mathrm{pred}}
    \boldsymbol\Sigma_{\mathrm{GP}}(\mathbf x)
  \right),
  \label{eq:prediction_posterior}\\
  p_{\mathrm{acq}}(\mathbf f(\mathbf x)
                   \mid\mathcal D_n,S_{n+1}=s)
  &=
  \mathcal N\!\left(
    \boldsymbol\mu_{\mathrm{GP}}(\mathbf x)
    +
    \mathbf s_n\odot\widehat{\mathbf b}_{s,n},
    c_{\mathrm{acq}}
    \boldsymbol\Sigma_{\mathrm{GP}}(\mathbf x)
  \right).
  \label{eq:app_decision_posterior}
\end{align}
The first is used for predictive diagnostics, whereas the second is the
decision posterior supplied to qLogEHVI.

We use \(W_b=30\), \(H_b=8\), \(\lambda_0=5\),
\(\lambda_s=3\), \(\kappa_b=2.5\), and \(\tau_{\min}=0.10\)
for mean correction. Spatial calibration uses \(G=5\), \(c_0=1\),
\(\lambda_{\mathrm{CV}}=20\), \(\omega=0.75\),
\(c_{\min}=0.25\), and \(c_{\max}^{\mathrm{CV}}=1000\).
Selected-point calibration uses \(W_c=12\), \(H_c=4\),
\(\lambda_c=2\), \(\kappa_c=2.5\), and
\(c_{\max}^{\mathrm{sel}}=64\). Finally,
\(\gamma=1/4\) and \(c_{\mathrm{cap}}=3\).

\subsection{Pareto-local and periodic-global qLogEHVI}
\label{app:local_global_details}

The fifth diagnostic in Appendix~\ref{app:pilot_diagnostics} concerns the
numerical optimization of the acquisition function. Full-domain multistart
optimization can send several restarts toward similar distant or
boundary-adjacent regions, where the nonstationary surrogate is weakly
supported. We retain standard qLogEHVI but organize most of its restarts around
diverse observed Pareto points. Periodic global rounds preserve exploration of
the complete domain. Importantly, the GP itself remains global and is fitted
to all observations; localization is applied only to acquisition optimization.

Let
\begin{equation}
  \mathcal I_n
  =
  \left\{
    i\in\{1,\ldots,n\}:
    \nexists j\text{ such that }\mathbf y_j
    \text{ dominates }\mathbf y_i
  \right\}
  \label{eq:observed_pareto_indices}
\end{equation}
index the currently observed nondominated points. For each decision coordinate
and objective, we apply min--max normalization over this set:
\begin{align}
  \widetilde x_{i,d}
  &=
  \frac{
    x_{i,d}-\min_{j\in\mathcal I_n}x_{j,d}
  }{
    \max\{
      \max_{j\in\mathcal I_n}x_{j,d}
      -
      \min_{j\in\mathcal I_n}x_{j,d},
      \varepsilon_z
    \}
  },
  \label{eq:pareto_x_normalization}\\
  \widetilde y_{i,m}
  &=
  \frac{
    y_{i,m}-\min_{j\in\mathcal I_n}y_{j,m}
  }{
    \max\{
      \max_{j\in\mathcal I_n}y_{j,m}
      -
      \min_{j\in\mathcal I_n}y_{j,m},
      \varepsilon_z
    \}
  }.
  \label{eq:pareto_y_normalization}
\end{align}
The center-selection feature is
\begin{equation}
  \mathbf z_i
  =
  \begin{bmatrix}
    \widetilde{\mathbf x}_i\\
    \widetilde{\mathbf y}_i
  \end{bmatrix},
  \qquad i\in\mathcal I_n.
  \label{eq:pareto_center_features}
\end{equation}

Let \(B\) denote the total acquisition-restart budget and \(K_{\max}\) the
maximum number of regions. The number of centers is
\begin{equation}
  K_n
  =
  \min\{K_{\max},B,|\mathcal I_n|\}.
  \label{eq:number_of_pareto_centers}
\end{equation}
If \(|\mathcal I_n|\leq K_n\), all observed nondominated points are retained.
Otherwise, the first center is the observed extreme maximizing the first
objective,
\begin{equation}
  i_1
  \in
  \arg\max_{i\in\mathcal I_n}y_{i,1},
  \label{eq:first_pareto_center}
\end{equation}
and subsequent centers are selected by farthest-point traversal:
\begin{equation}
  i_k
  \in
  \arg\max_{
    i\in\mathcal I_n\setminus\{i_1,\ldots,i_{k-1}\}
  }
  \min_{\ell<k}
  \frac{1}{D+M}
  \left\|
    \mathbf z_i-\mathbf z_{i_\ell}
  \right\|_2^2,
  \qquad k=2,\ldots,K_n.
  \label{eq:pareto_center_selection}
\end{equation}

Around center \(\mathbf x_{i_k}\), we define the axis-aligned region
\begin{equation}
  \mathcal R_{n,k}(r_n)
  =
  \left[
    \max\{
      \underline{\mathbf x},
      \mathbf x_{i_k}
      -
      r_n(\overline{\mathbf x}-\underline{\mathbf x})
    \},
    \;
    \min\{
      \overline{\mathbf x},
      \mathbf x_{i_k}
      +
      r_n(\overline{\mathbf x}-\underline{\mathbf x})
    \}
  \right].
  \label{eq:pareto_region}
\end{equation}
Thus, \(r_n\) is a half-width expressed as a fraction of the full range of
each decision coordinate.

The \(B\) restarts are allocated as evenly as possible:
\begin{equation}
  B_{n,k}
  =
  \left\lfloor\frac{B}{K_n}\right\rfloor
  +
  \mathbb I
  \left\{
    k\leq B\bmod K_n
  \right\},
  \qquad
  \sum_{k=1}^{K_n}B_{n,k}=B.
  \label{eq:restart_allocation}
\end{equation}
For region \(k\), the raw local initialization pool has size
\begin{equation}
  N_{n,k}^{\mathrm{raw}}
  =
  \max\!\left\{
    16B_{n,k},
    \left\lceil
      N_{\mathrm{raw}}\frac{B_{n,k}}{B}
    \right\rceil
  \right\}.
  \label{eq:local_raw_pool_size}
\end{equation}
Its points are sampled around the center and clipped to the region:
\begin{equation}
  \mathbf u_{n,k,j}
  =
  \Pi_{\mathcal R_{n,k}(r_n)}
  \left[
    \mathbf x_{i_k}
    +
    \sigma_n
    (\overline{\mathbf x}-\underline{\mathbf x})
    \odot\boldsymbol\xi_j
  \right],
  \qquad
  \boldsymbol\xi_j\sim\mathcal N(\mathbf 0,\mathbf I),
  \qquad
  \sigma_n=\min\{\sigma_0,r_n/2\}.
  \label{eq:local_restart_pool}
\end{equation}
The \(B_{n,k}\) pool points with the largest acquisition values initialize
bounded gradient optimization inside \(\mathcal R_{n,k}(r_n)\).

Let \(u=1,\ldots,T\) index the BO evaluations after an initial design of
size \(n_0\). Before iteration \(u\), \(n=n_0+u-1\) evaluations have been
completed. The search mode for the next candidate \(\mathbf x_{n+1}\) is
\begin{equation}
  S_{n+1}
  =
  \begin{cases}
    \mathrm{global},
      & u\leq40\ \text{and}\ u\bmod5=0,\\
    \mathrm{global},
      & u>40\ \text{and}\ u\bmod10=0,\\
    \mathrm{local},
      & \text{otherwise}.
  \end{cases}
  \label{eq:search_schedule}
\end{equation}
Hence, global rounds constitute \(20\%\) of the first 40 post-initial-design
iterations and \(10\%\) thereafter.

For each mode \(s\in\{\mathrm{local},\mathrm{global}\}\), define the acquisition
function using the corresponding decision posterior in
Eq.~\eqref{eq:decision_posterior}:
\begin{equation}
  \alpha_n(\mathbf x;s)
  =
  \operatorname{qLogEHVI}
  \left(
    \mathbf x;
    p_{\mathrm{acq}}(
      \mathbf f(\mathbf x)
      \mid\mathcal D_n,S_{n+1}=s)
  \right).
  \label{eq:local_global_qlogehvi}
\end{equation}
On a local round, let \(\mathbf x_{n,k,b}^{\star}\) denote the result of
restart \(b\) optimized within region \(k\). We select the best
region--restart pair and return its candidate:
\begin{equation}
  \begin{aligned}
    (k^\star,b^\star)
    &\in
    \arg\max_{
      \substack{
        1\leq k\leq K_n\\
        1\leq b\leq B_{n,k}
      }
    }
    \alpha_n(\mathbf x_{n,k,b}^{\star};\mathrm{local}),\\
    \mathbf x_{n+1}
    &=
    \mathbf x_{n,k^\star,b^\star}^{\star}.
  \end{aligned}
  \label{eq:local_candidate_selection}
\end{equation}
On a global round, a scrambled Sobol pool of size
\begin{equation}
  N_{\mathrm{global}}^{\mathrm{raw}}
  =
  \max\{N_{\mathrm{raw}},16B\}
  \label{eq:global_raw_pool_size}
\end{equation}
is generated over \(\mathcal X\). Its \(B\) highest-valued points initialize
full-domain optimization. Let \(\mathbf x_{n,b}^{\star}\in\mathcal X\)
denote the result of restart \(b\). We select the best restart and return
its candidate:
\begin{equation}
  \begin{aligned}
    b^\star
    &\in
    \arg\max_{1\leq b\leq B}
    \alpha_n(\mathbf x_{n,b}^{\star};\mathrm{global}),\\
    \mathbf x_{n+1}
    &=
    \mathbf x_{n,b^\star}^{\star}.
  \end{aligned}
  \label{eq:global_candidate_selection}
\end{equation}

\paragraph{Adaptive local radius.}
After evaluating a local candidate \(\mathbf x_{n+1}\), define the realized
hypervolume improvement
\begin{equation}
  \Delta\mathrm{HV}_{n+1}
  =
  \mathrm{HV}(\mathcal D_{n+1})
  -
  \mathrm{HV}(\mathcal D_n).
  \label{eq:realized_hv_improvement}
\end{equation}
A local round is considered successful if
\(\Delta\mathrm{HV}_{n+1}>\epsilon_{\mathrm{HV}}\).
On local rounds, the common radius is updated as
\begin{equation}
  r_{n+1}
  =
  \begin{cases}
    \min\{r_{\max},\eta_{\mathrm{inc}}r_n\},
      & \text{after three consecutive local successes},\\
    \max\{r_{\min},\eta_{\mathrm{dec}}r_n\},
      & \text{after five consecutive local failures},\\
    r_n,
      & \text{otherwise}.
  \end{cases}
  \label{eq:radius_update}
\end{equation}
The corresponding counter is reset whenever the other outcome occurs and
after a radius expansion or contraction. Global rounds satisfy
\(r_{n+1}=r_n\) and leave both local counters unchanged.

We use \(B=8\), \(N_{\mathrm{raw}}=128\), \(K_{\max}=6\),
\(r_{n_0}=0.20\), \(r_{\min}=0.10\), \(r_{\max}=0.25\),
\(\sigma_0=0.10\), \(\epsilon_{\mathrm{HV}}=10^{-4}\),
\(\eta_{\mathrm{inc}}=1.2\), and \(\eta_{\mathrm{dec}}=0.8\).

\paragraph{Computational cost.}
With $n$ complete observations of $M$ objectives, dense exact inference
uses an $nM\times nM$ covariance matrix. A Cholesky factorization requires
$\mathcal O((nM)^3)$ time and $\mathcal O((nM)^2)$ storage, with repeated
factorizations during hyperparameter optimization. Fixed-hyperparameter
calibration uses the fitted covariance and precision submatrices without
refitting the GP for each fold. Sequential fits are warm-started, and
local regions share a fixed acquisition-restart budget. Localization
changes the acquisition domain but not the GP training set. The reported
benchmarks contain at most 110 observations of two objectives, corresponding
to 220 scalar observations.

\section{Additional performance and robustness analyses}
\label{sec:additional_results}

\paragraph{Run-to-run robustness and alternative Pareto-quality metrics.}
We first complement the mean-performance results with adverse-tail statistics
computed across seeds within each problem. For nHV, the adverse tail is its
10th percentile, whereas for nIGD and the two AUC measures it is the 90th
percentile. Table~\ref{tab:additional_metric_ranks} reports the corresponding
mean problem ranks. MOLRN-BO ranks first on all four tail measures and also has
the smallest mean across-seed standard deviation of final nIGD among the nine
algorithms ($0.0358$). These results indicate that its main robustness benefit is a reduced incidence
of poorly performing trials, rather than a uniform reduction in variance on
every problem and metric.

We additionally evaluate two distance-based indicators in the ideal--nadir
normalized objective space. For an approximation set $A$ and reference front
$R$, normalized IGD$^+$ is defined as
\begin{equation}
\operatorname{nIGD}^{+}(A,R)
=
\frac{1}{|R|}
\sum_{\mathbf r\in R}
\min_{\mathbf a\in A}
\left\|
\max(\mathbf r-\mathbf a,\mathbf 0)
\right\|_2,
\end{equation}
and the normalized additive $\epsilon$-indicator is
\begin{equation}
I_{\epsilon+}(A,R)
=
\max_{\mathbf r\in R}
\min_{\mathbf a\in A}
\max_j(r_j-a_j).
\end{equation}
Both indicators are lower-is-better. MOLRN-BO obtains the best mean problem
rank for nIGD$^+$ ($2.25$) and lies in the top three on $11/12$ problems.
It outperforms the original MO-LRN on all 12 problems under this metric.
For the additive $\epsilon$-indicator, qLogNParEGO ranks first ($2.83$) and
MOLRN-BO ranks second ($3.17$). MOLRN-BO is better than qLogNParEGO on only
$5/12$ problems, although it remains better than the original MO-LRN on
$10/12$. This difference is consistent with the indicators' emphases:
nIGD$^+$ averages directional shortfalls over the reference front, whereas
$I_{\epsilon+}$ is controlled by the most poorly covered reference point.
Thus, MOLRN-BO provides strong average front coverage, while qLogNParEGO is
slightly stronger in worst-location coverage.

\begin{table*}[t]
  \centering
  \scriptsize
  \setlength{\tabcolsep}{4pt}
  \caption{Mean problem ranks for adverse-tail and additional Pareto-quality
  metrics over 12 benchmarks using all valid trials. Lower is better.
  Tail quantiles are computed across seeds within each problem. Bold and
  underlined entries denote the best and second-best ranks, respectively.}
  \label{tab:additional_metric_ranks}
  \begin{tabular}{lcccccc}
    \toprule
    Method
    & nHV-q10
    & nIGD-q90
    & HV-regret-AUC-q90
    & nIGD-AUC-q90
    & Final nIGD$^+$
    & Final $I_{\epsilon+}$ \\
    \midrule
    MOLRN-BO
    & \textbf{2.08}
    & \textbf{2.08}
    & \textbf{2.00}
    & \textbf{2.83}
    & \textbf{2.25}
    & \underline{3.17} \\
    qLogNParEGO
    & \underline{2.83}
    & \underline{3.50}
    & \underline{3.00}
    & \underline{3.33}
    & \underline{2.50}
    & \textbf{2.83} \\
    qLogEHVI
    & 4.08 & 4.83 & 3.58 & 3.92 & 4.00 & 4.67 \\
    mobopt
    & 5.12 & 4.33 & 5.54 & 4.67 & 4.58 & 3.67 \\
    JES
    & 5.21 & 4.58 & 5.21 & 4.25 & 4.75 & 4.42 \\
    Original MO-LRN
    & 6.12 & 5.08 & 6.29 & 5.75 & 5.83 & 5.33 \\
    NSGA-II
    & 5.62 & 6.50 & 6.29 & 7.17 & 6.33 & 7.50 \\
    MESMO
    & 7.21 & 7.00 & 7.12 & 6.50 & 7.00 & 5.75 \\
    Sobol
    & 6.71 & 7.08 & 5.96 & 6.58 & 7.75 & 7.67 \\
    \bottomrule
  \end{tabular}
\end{table*}

\paragraph{Convergence behavior.}
Figure~\ref{fig:performance_trajectories} reports the nHV and nIGD trajectories
over the complete evaluation budget. MOLRN-BO's advantage is generally more
pronounced late in optimization. Across the nine algorithms, its mean nIGD
problem rank improves from $3.33$ after 25 post-initial evaluations to $2.67$,
$2.42$, and $2.17$ after 50, 75, and 100 evaluations, respectively. Its nHV
rank improves more gradually from $2.58$ at 25 evaluations to $2.25$ at the
final budget. This late improvement explains why MOLRN-BO achieves the best
final nHV and nIGD ranks while qLogNParEGO remains slightly better on the two
anytime AUC measures.

The trajectories also reveal meaningful problem dependence. MOLRN-BO is
comparatively slow on WFG1, performs weakly in both final nHV and anytime
performance on WFG5, and exhibits an extended nIGD plateau on the disconnected
DTLZ7 front. In contrast, its final coverage improves substantially on many of
the remaining ten-dimensional problems and is consistently better than that
of the original MO-LRN. The curves therefore support a conclusion of strong
final quality and reduced poor-trial risk, but not uniformly faster convergence
over the entire benchmark suite.

\end{document}